\documentclass[twoside]{article}
\usepackage[preprint]{AISTATS2026PaperPack/aistats2026}\usepackage{natbib}

\usepackage[utf8]{inputenc} 
\usepackage[T1]{fontenc}    
\usepackage{lmodern}
\usepackage{hyperref}       
\usepackage{url}            
\usepackage{booktabs}       
\usepackage{amsfonts}       
\usepackage{nicefrac}       
\usepackage{microtype}      

\usepackage[linesnumbered, ruled, vlined, noend]{algorithm2e}
\usepackage{amsmath}
\usepackage{amsthm}
\usepackage{amssymb}

\usepackage{bm}
\usepackage[title]{appendix}
\usepackage{courier}
\usepackage{enumitem}
\usepackage{graphicx}
\usepackage{algorithmic}
\usepackage[caption=false,font=footnotesize]{subfig}
\usepackage[margin=1in]{geometry}
\usepackage{authblk}
\usepackage{nicefrac}
\usepackage{mathtools}
\usepackage{color}
\usepackage{xspace} 
\usepackage{times}
\usepackage{hyperref}
\usepackage{stmaryrd}

\newcommand{\Rbb}{\mathbb{R}}
\newcommand{\R}{\mathbb{R}}
\newcommand{\pinv}{+}

\usepackage{todonotes}

\newtheorem*{theorem*}{Theorem}
\newtheorem*{lemma*}{Lemma}

\RequirePackage{latexsym}
\RequirePackage{amsmath}
\RequirePackage{amssymb}
\RequirePackage{bm}
\RequirePackage[mathscr]{euscript}

\newcommand{\V}{\Vb}

\newcommand{\ts}{\top}

\newcommand{\RR}[2]{\mathbb{R}^{#1 \times #2}} 

\newcommand{\ab}{\mathbf{a}}
\newcommand{\bb}{\mathbf{b}}

\newcommand{\qb}{\mathbf{q}}

\newcommand{\Ab}{\mathbf{A}}
\newcommand{\Bb}{\mathbf{B}}

\newcommand{\Hb}{\mathbf{H}}
\newcommand{\Ib}{\mathbf{I}}

\newcommand{\Kb}{\mathbf{K}}

\newcommand{\Mb}{\mathbf{M}}

\newcommand{\Pb}{\mathbf{P}}
\newcommand{\Qb}{\mathbf{Q}}

\newcommand{\Sb}{\mathbf{S}}

\newcommand{\Ub}{\mathbf{U}}
\newcommand{\Vb}{\mathbf{V}}
\newcommand{\Wb}{\mathbf{W}}
\newcommand{\Xb}{\mathbf{X}}

\renewcommand{\vec}[1]{\mathbf{\boldsymbol{#1}}}

\newcommand{\vv}{\vec{v}}

\ifx\BlackBox\undefined
\newcommand{\BlackBox}{\rule{1.5ex}{1.5ex}}  
\fi

\ifx\QED\undefined
\def\QED{~\rule[-1pt]{5pt}{5pt}\par\medskip}
\fi

\ifx\proof\undefined
\newenvironment{proof}{\par\noindent{\bf Proof\ }}{\hfill\BlackBox\\[2mm]}
\fi

\ifx\theorem\undefined
\newtheorem{theorem}{Theorem}
\fi

\ifx\example\undefined
\newtheorem{example}{Example}
\fi

\ifx\property\undefined
\newtheorem{property}{Property}
\fi

\ifx\lemma\undefined
\newtheorem{lemma}[theorem]{Lemma}
\fi

\ifx\proposition\undefined
\newtheorem{proposition}[theorem]{Proposition}
\fi

\ifx\remark\undefined
\newtheorem{remark}[theorem]{Remark}
\fi

\ifx\corollary\undefined
\newtheorem{corollary}[theorem]{Corollary}
\fi

\ifx\definition\undefined
\newtheorem{definition}{Definition}
\fi

\ifx\step\undefined
\newtheorem{step}{Step}
\fi

\ifx\conjecture\undefined
\newtheorem{conjecture}[theorem]{Conjecture}
\fi

\ifx\axiom\undefined
\newtheorem{axiom}[theorem]{Axiom}
\fi

\ifx\claim\undefined
\newtheorem{claim}[theorem]{Claim}
\fi

\ifx\assumption\undefined
\newtheorem{assumption}[theorem]{Assumption}
\fi

\renewcommand{\RR}{\mathbb{R}} 

\newcommand{\norm}[1]{\|#1\|}

\newcommand{\Sigmab}{\mathbf{\Sigma}}

\newcommand{\Km}{\ensuremath\mathbf{K}}
\newcommand{\Qm}{\ensuremath\mathbf{Q}}

\newcommand{\Vm}{\ensuremath\mathbf{V}}

\newcommand{\Wm}{\ensuremath\mathbf{W}}
\newcommand{\Xm}{\ensuremath\mathbf{X}}

\newcommand{\hv}{\ensuremath\mathbf{h}}
\newcommand{\kv}{\ensuremath\mathbf{k}}
\newcommand{\qv}{\ensuremath\mathbf{q}}
\newcommand{\xv}{\ensuremath\mathbf{x}}

\newcommand{\Q}{\Qb}
\newcommand{\K}{\Kb}
\newcommand{\A}{\Ab}
\newcommand{\B}{\Bb}

\usepackage{xspace}
\newcommand{\kqsvd}{KQ-SVD\xspace}
\newcommand{\methodname}{KQ-SVD\xspace}
\newcommand{\ksvd}{K-SVD\xspace}
\newcommand{\eigen}{Eigen\xspace}
\newcommand{\gqa}{GQA\xspace}
\newcommand{\Softmax}{\text{Softmax}}

\begin{document}

%

%

\runningtitle{KQ-SVD: Compressing the KV Cache}

\twocolumn[

\aistatstitle{KQ-SVD: Compressing the KV Cache with\\ Provable Guarantees on Attention Fidelity}

\aistatsauthor{ Damien Lesens \And Beheshteh T. Rakhshan \And  Guillaume Rabusseau}

\aistatsaddress{ ENS de Lyon \And DIRO,
Université de Montréal\\ Mila \And DIRO,
Université de Montréal\\ Mila - CIFAR AI Chair} ]

\begin{abstract}
The Key–Value (KV) cache is central to the efficiency of transformer-based large language models (LLMs), storing previously computed vectors to accelerate inference. Yet, as sequence length and batch size grow, the cache becomes a major memory bottleneck. Prior compression methods typically apply low-rank decomposition to keys alone or attempt to jointly embed queries and keys, but both approaches neglect that attention fundamentally depends on their inner products. In this work, we prove that such strategies are sub-optimal for approximating the attention matrix. We introduce \methodname, a simple and computationally efficient method that directly performs an optimal low-rank decomposition of the attention matrix via a closed-form solution. By targeting the true source of redundancy, \methodname~preserves attention outputs with higher fidelity under compression. Extensive evaluations on LLaMA and Mistral models demonstrate that our approach consistently delivers superior projection quality.
\end{abstract}

\section{Introduction}
The rise of Large Language Models (LLMs)~\citep{touvron2023llama, chaplot2023albert, achiam2023gpt, guo2025deepseek} has expanded AI capabilities beyond earlier models. Transformers~\citep{vaswani2017attention} replace recurrence with self-attention, enabling parallelism and improved sequence modeling, but their quadratic memory and computation costs limit long-sequence scalability. 

Key-Value (KV) caches are introduced to accelerate autoregressive generation by storing intermediate attention KV vectors, avoiding redundant computation of shared prefixes for each generated token. Although KV caching reduces computational overhead, it substantially increases memory consumption, as the cache size grows linearly with both sequence length and batch size. This trade-off motivates the development of KV cache compression techniques, which are crucial for enabling efficient and cost-effective deployment of LLMs across diverse hardware platforms~\citep{fu2024challenges, shi2024keep}.
Variants like Multi-Query Attention (MQA)~\citep{shazeer2019fast} and Grouped-Query Attention (GQA)~\citep{ainslie2023GQA} reduce KV cache size by sharing or grouping query vectors while maintaining performance comparable to full Multi-Head Attention~(MHA). However, they may introduce accuracy trade-offs and hardware sensitivity, which can affect performance generalization. Additional approaches, including sparse~\citep{zhang2021sparse} and linearized~\citep{katharopoulos2020transformers} attention, further reduce computational and memory costs, shaping KV cache optimization strategies.

Another promising line of research exploits the low-rank structure of KV caches to reduce memory overhead. Multi-Head Latent Attention~(MLA)~\citep{liu2024deepseek, guo2025deepseek} maps tokens into the low-rank latent space and stores these compressed representations in place of the original key and value states. However, using MLA necessitates training the model from the ground up. In contrast, ASVD~\citep{yuan2024asvd}, LoRC~\citep{zhang2024lorc}, and Palue~\citep{changpalu} apply SVD to key-value parameter matrices without retraining to build low-rank projection modules. A key limitation of these approaches is that they often compress only the keys, neglecting the query-key interaction that underlies attention. EigenAttention~\citep{saxena2024eigen} and Zack~\citep{zhang2024zack} attempt to address this by incorporating both queries and keys in low-rank decompositions, yet their behavior largely resembles that of SVD-based methods that compress keys alone.

In this work, we address these limitations by introducing \kqsvd, a compression method that achieves optimal low-rank approximation of the attention matrix efficiently and in closed form. Our method explicitly captures the interactions between queries and keys through their inner products, preserving the fundamental structure of attention. Beyond key-query interactions, we also consider the corresponding interactions between values and the output projection, enhancing the fidelity of the approximation. By leveraging the inherent low-rank structure of KV caches~\citep{yu2024effectively, saxena2024eigen}, we formulate attention matrix approximation as a principled low-rank decomposition problem. Our theoretical analysis quantifies the error between prior key-only SVD approaches and our optimal method, and shows that methods incorporating both queries and keys can degrade when keys and queries are rescaled by the same factor, effectively behaving like key-only SVD methods.
Our contributions can be summarized as follows:
\begin{itemize}
    \item We introduce \kqsvd, an optimal low-rank approximation of the attention matrix capturing key-query interactions.
    \item We theoretically quantify the advantages of \methodname over methods based on key low-rank decomposition and SVD on concatenated queries and keys.
    \item  We show that \methodname is compatible with and also optimal in the Grouped-Query Attention setting.
    \item We provide  extensive empirical evaluations with LLaMA2-7B,  LLaMA2-13B, LLama3-8B and Mistral-7B models on the C4 dataset demonstrating significant advantages of \methodname over existing low rank projection methods.
\end{itemize}

\section{Related Works}
\paragraph{Low-rank structure of the KV-cache.}
Several methods exploit the inherent low-rank structure of cached key–value (KV) matrices to reduce memory footprint. ECKVH~\citep{yu2024effectively} compresses the cache by grouping attention heads, performing singular value decomposition (SVD) within each group, and retaining only the dominant singular components. EigenAttention~\citep{saxena2024eigen} generalizes this idea by constructing low-rank bases that jointly approximate queries, keys, and values, effectively lowering the dimensionality of KV representations. Q-Filters~\citep{godeyq} introduces a training-free variant, projecting keys into a low-rank subspace via SVD to approximate attention scores efficiently with minimal accuracy loss. Moreover,~\citep{yu2024effectively} investigates the intrinsic low-rank nature of KV caches and compresses KV heads through careful grouping and SVD-based decomposition. In contrast, Loki~\citep{singhania2024loki} adopts a two-stage strategy: it first estimates approximate attention scores in a lower-dimensional space to rank and select the most relevant keys, and then computes exact attention scores using only the selected keys, reducing both memory and computational cost.
\paragraph{KV Weights Compression.}
An alternative approach targets the KV weight matrices themselves rather than the cached matrices. LoRC~\citep{zhang2024lorc} applies low-rank approximations directly to the key and value weight matrices, achieving compression at the parameter level. Palu~\citep{changpalu} follows a similar strategy, jointly compressing key and value weight matrices via SVD. ShadowKV~\citep{sun2024shadowkv} introduces a distinct perspective by performing SVD on pre-RoPE key matrices to reduce their dimensionality, demonstrating the versatility of low-rank methods in optimizing KV representations.
\paragraph{Positioning of \kqsvd.}
Although prior methods have made significant strides in compressing KV caches and their weight matrices, they often treat keys and values independently or approximate attention only indirectly, leaving the core query–key interactions underrepresented and leading to sub-optimal low rank approximation of attention matrices. \kqsvd addresses these limitations by formulating a principled, closed-form low-rank approximation of the full attention matrix.
\section{Preliminary}
\label{sec:prelim}
In this section, we introduce our notations and present the
necessary background on Multi-Head Attention~(MHA). 
\subsection{Notations}
We use lower case bold letters for vectors~(e.g.,   $\ab,\bb$), upper case bold letters for matrices~(e.g., $\Ab,\Bb$). $\Ab^{+}$ denotes the Moore–Penrose pseudo-inverse of $\Ab$. Throughout the paper, the singular value decomposition (SVD) of a matrix $\Sb\in\RR^{m\times n}$ is presented by $\Sb=\Ub \Sigmab \Vb^\ts$, with $\Ub \in\R^{m\times n}$, $\Vb\in \RR^{n\times n}$ matrices with orthonormal columns and $\Sigmab \in \RR^{n\times n}$ a diagonal matrix with positive diagonal entries $\{\sigma_i \}_{i=1}^n$. The columns of $\Ub$ and $\Vb$ are called respectively the left and right singular vectors of $\Sb$, and the $\sigma_i$'s the singular values of $\Sb$, noted $\sigma_i(\Sb)$. 
The optimal rank-$R$ approximation of $\Sb$ with respect to the Frobenius norm can be obtained via the SVD by truncating it to keep only the first $R$ singular vectors and singular values: ${\Sb} \approx \hat{\Ub} \hat{\Sigmab} \hat{\Vb}^\ts$ with $\hat{\Ub} \in \RR^{m \times R}$, $\hat{\Vb} \in \RR^{n \times R}$  and $\hat{\Sigmab} \in \RR^{R \times R}$. 
The column space of $\Sb$ is noted $\mathcal{R}(\Sb)$.
%
\subsection{Background}\label{sec:background}
In transformer architectures, self-attention assigns relative importance to tokens, enabling the model to selectively focus on different segments of the input sequence.  
For a sequence of token embeddings $\Xb \in \RR^{T \times D}$, multi-head attention is computed as
$$
\mathrm{MHA}(\Xb) = [\Hb_1, \dots, \Hb_h] \Wb^O,
$$
where
$$
\Hb_i = \mathrm{Softmax}\Big(\frac{\Qb_i \Kb_i^\ts}{\sqrt{d}}\Big) \Vb_i,
$$
with $\Wb_i^Q, \Wb_i^K, \Wb_i^V \in \RR^{D \times d}$, $d = D/h$, $\Qb_i = \Xb \Wb_i^Q$, $\Kb_i = \Xb \Wb_i^K$, $\Vb_i = \Xb \Wb_i^V$, and 
 $\Wb^O \in \RR^{D \times D}$.  
In masked attention, the upper-diagonal entries of the attention matrix $\Qb_i \Kb_i^\ts$ are set to $-\infty$ to prevent a token from attending to future positions.

This computation scales quadratically with the sequence length $T$. In auto-regressive decoding, previously computed key and value vectors are cached to avoid redundant computation, reducing the per-token cost. Specifically at time $T$, for each head, new key–value pairs are concatenated to the caches $\Km$ and $\Vm$, followed by an attention computation:
\begin{align*}
    \Km \leftarrow &\text{Concat}(\Km, \kv_T),\
    \Vm \leftarrow \text{Concat}(\Vm, \vv_T), \\
    &\hv_T = \text{Softmax}\Big(\frac{\qv_T \Km^\ts}{\sqrt{d}}\Big) \Vm
\end{align*}
where $\kv_T = \xv_T\Wb^K$, $\vv_T = \xv_T\Wb^V$, $\qv_T = \xv_T\Wb^Q$.
While caching mitigates redundant computation, generating the $T$-th token still incurs $\mathcal{O}(T)$ cost, and the memory footprint of stored keys and values grows linearly with sequence length. For sufficiently long contexts, this memory requirement becomes a dominant bottleneck, as the cumulative size of the KV cache can surpass the model parameters by several orders of magnitude.
In the following sections, we briefly review the \ksvd~method~\citep{changpalu, yu2024effectively, zhang2024lorc}, which compresses key representations using singular value decomposition (SVD), and the \eigen~approach~\citep{saxena2024eigen}, which jointly considers keys and queries by vertically concatenating them and applying SVD to the resulting matrix.
\subsection{Cache compression with SVD}\label{sec:ksvd}

Recent works~\citep{changpalu,zhang2024lorc,chang2025xkv} demonstrate that singular value decomposition (SVD) is a powerful tool for compressing the KV cache in large language models, as the cache exhibits low-rank structure.
Let $\Kb = \Ub_K \Sigmab_K \Vb_K^\ts \in \R^{T\times d}$ be the SVD of the key matrix, and let $\widetilde{\Kb} = \hat{\Ub}_K \hat{\Sigmab}_K  \hat{\Vb}_K^\ts$ denote its rank-$R$ truncated version. 
By the Eckart–Young–Mirsky theorem, $\widetilde{\Kb}$ is the best rank-$R$ approximation of $\Kb$ under the Frobenius norm. In other words, the optimization problem
$$\min_{\Pb \in \R^{d\times d}} \|\Kb \Pb - \Kb\|_F^2 \quad \text{s.t.} \quad \mathrm{rank}(\Pb)\le R,$$
is solved by $\Pb = \hat{\Vb}_K\hat{\Vb}_K^\ts$, leading to the approximation
$\widetilde{\Kb} = \Kb  \hat{\Vb}_K\hat{\Vb}_K^\ts = \hat{\Ub}_K\hat{\Sigmab}_K\hat{\Vb}_K^\ts$.
Applying the same procedure to the value matrix $\Vb = \Ub_V \Sigmab_V \Vb_V^\ts$, we can approximate the attention output as
\begin{align*}
    \widetilde{\Hb} 
    &= \mathrm{Softmax}(\Qb \widetilde{\Kb}^\ts/ \sqrt{d}) \widetilde{\Vb} \\
    &= 
    \mathrm{Softmax}(\Qb \hat{\Vb}_K\hat{\Vb}_K^\ts \Kb^\ts/ \sqrt{d}) \Vb \hat{\Vb}_V \hat{\Vb}_V^\ts.
\end{align*}
This formulation is particularly useful because it allows to store only the compressed caches $\Kb \hat{\Vb}_K$ and $\Vb \hat{\Vb}_V$ in memory. These matrices are of size $R \times T$ instead of $d \times T$, resulting in significant memory savings. At runtime, queries are multiplied by $\hat{\Vb}_K$, while $\hat{\Vb}_V^\ts$ can be absorbed into the output projection $\Wb_O$, streamlining computation.

A key advantage of this approach is that the SVD does not need to be computed during token generation. Instead, it can be performed once in a post-training calibration phase. For each layer $l$ and attention head $i$, we only need to determine a basis $\Vb_{i,l} \in \R^{d \times R}$ such that $\Km_{i,l} \approx \Km_{i,l} \Vb_{i,l} (\Vb_{i,l})^\ts$
with an analogous construction for the values. This basis can be learned from a calibration set of sequences. Specifically, we pass $n_s$ calibration sequences (e.g., sampled from a high-quality dataset such as C4~\citep{C4}) through the model. The $k$th sequence produces caches $\Kb_{i,l}^k$ and $\Vb_{i,l}^k$ for every layer $l$ and head $i$. These are concatenated to form large cache matrices 
$\Kb_{i,l} = \begin{pmatrix}
\Kb_{i,l}^1, \Kb_{i,l}^2, \dots, \Kb_{i,l}^{n_s}
\end{pmatrix}$.
This aggregated cache provides a representative sample of the key vectors that will appear during inference. Performing SVD on $\Kb_{i,l}$ then yields the dominant singular vectors, which form a suitable low-rank basis for compression. The cost of generating calibration caches and computing the SVDs is negligible compared to model training, and is offset by the runtime speedups from cache compression. In the following, we will refer to this method as \ksvd.

\textbf{Rank selection.} The compression rank $R$ is determined per layer by examining the singular value spectrum. Let $\{\sigma_j\}_j$ denote the singular values of a matrix $\Mb$. For a relative error tolerance $\epsilon$, we select the smallest $R$ such that
$$\|\Mb - \widetilde{\Mb}\|_F^2 \le \epsilon \|\Mb\|_F^2  \Leftrightarrow \frac{\sum_{j=1}^{R} \sigma_j^2}{\sum_{i=1}^d \sigma_j^2} \ge 1 -\epsilon.$$
Prior studies~\citep{yu2024effectively,saxena2024eigen} have shown that KV matrices are indeed approximately low-rank, so substantial compression can be achieved with small error budget $\epsilon$. 
The chosen rank may differ for keys and values depending on their spectra.

\subsection{Cache compression with Eigen}\label{sec:eigen}
Other works~\citep{saxena2024eigen, zhang2024zack} emphasize that queries should also be considered when compressing key caches. Indeed, by projecting the key cache we also project the query matrix: $\Qb \hat{\Vb}_K \hat{\Vb}_K^\ts \Kb^\ts = (\Qb \hat{\Vb}_K \hat{\Vb}_K^\ts) (\Kb \hat{\Vb}_K\hat{\Vb}_K^\ts)^\ts$ as $\hat{\Vb}_K\hat{\Vb}_K^\ts$ is an idempotent matrix. Hence, it makes sense to compute the low rank projection by solving 
$$
\min_{\Sb} \| \Kb - \Kb \Sb\|_F^2 + \| \Qb - \Qb \Sb\|_F^2 \quad \text{s.t.} \quad \mathrm{rank}(\Sb) \leq R, $$
which is equivalent to
$$\min_{\Sb} \left\| 
\begin{bmatrix}
\Kb \\ 
\Qb
\end{bmatrix} - 
\begin{bmatrix}
\Kb \\ 
\Qb
\end{bmatrix} \Sb 
\right\|_F^2 
\quad \text{s.t.} \quad \mathrm{rank}(\Sb) \leq R,
$$
so that $\Sb$ will approximate queries and keys simultaneously. As the second formulation of the optimization problem shows, $\Sb$ can be computed by performing an SVD on the combined matrix $
\begin{bmatrix}
\Kb \\ 
\Qb
\end{bmatrix}
$. This approach ensures that the learned projection preserves keys and queries while reducing dimensionality. We refer to this approach as  \eigen  throughout this paper. The calibration process follows the same procedure as \ksvd: large calibration caches are formed by using a collection of calibration sequences. The only difference is that query matrices from the calibration set are also used in the projection computation.
\section{Methodology}
\label{sec:methodology}
In this section, we introduce our proposed approach \kqsvd~for KV cache compression. We consider taking the interaction between queries and keys into account. The method views the key and query matrices as a single entity and applies singular value decomposition~(SVD) to $\Km \Qm^\ts$. We begin by outlining the motivation behind this idea, followed by a detailed explanation of the technique.

\subsection{Motivation}
Existing compression methods based on SVD typically compress keys or jointly embed $\Qb/\Kb/\Vb$. 
Theorem~\ref{thm:kq-bound} (proof in Appendix \ref{apdx:thm1}) inspired by~\citep{wang2025squat} shows why that can fail: perturbations in $\Km$ are amplified by the inner products $\Qm \Km^\ts$ and further propagated by the value multiplication.

\begin{theorem}\label{thm:kq-bound}
     Let $\Xb\in\RR^{T\times d}$ be a sequence of token embeddings, $\Kb,\Qb, \Vb\in\RR^{T\times d}$ and 
     $$\mathrm{MHA}(\Xb) = \big[\mathrm{Softmax}(\Qb_i \Kb_i^\ts / \sqrt{d}) \Vb_i)\big]_i\Wb^{O},$$
    $$\widetilde{\mathrm{MHA}}(\Xb) = \big[\mathrm{Softmax}(\Qb_i \widetilde{\Kb_i}^\ts / \sqrt{d}) \widetilde{\Vb_i})\big]_i\Wb^{O},$$
where $\widetilde{\mathrm{MHA}}(\Xb), \widetilde{\Kb_i}$ and $\widetilde{\Vb_i}$ represent the approximation of $\mathrm{MHA}(\Xb),\Kb_i$ and $\Vb_i$, respectively. The difference between the actual attention output and the one produced with approximate keys and values is upper bounded as
\begin{align}\label{eq:purturbation-bound}
    &\norm{\widetilde{\mathrm{MHA}}(\Xm) - \mathrm{MHA}(\Xm)}_2\nonumber\\
    &\le 
    \sum_{i=1}^{h} \frac{\norm{\Vb_i \Wb_{i}^O}_2}{\sqrt{d}}\norm{\Qb_i \Kb_i^\ts - \Qb_i\widetilde{\Kb_i}^\ts}_2\nonumber\\
    &+ 
    \norm{\Vb_i\Wb_{i}^O-\widetilde{\Vb_i}\Wb_{i}^O }_2.
\end{align}
\end{theorem}

The method proposed by this paper stems from the will to minimize this upper bound on the output approximation error.
In each attention head, for a key cache $\Km$ and a calibration set of query vectors $\{\qb^{(j)}\}_{j\in\mathrm{calibration}}$ (row vectors of size $d$), we want to minimize 
$$\sum_{j\in\mathrm{calibration}}\norm{\qb^{(j)}\Kb^\ts-\qb^{(j)}\widetilde{\Kb}^\ts}_2^2,$$
over low-rank $\widetilde{\Kb}$. This objective is equivalent to a low-rank approximation of the query–key interaction matrix and admits an optimal closed-form solution via the singular value decomposition. Similarly, to compress values we directly optimize the second term of  Eq.~\eqref{eq:purturbation-bound} over low-rank $\widetilde{\Vm}$ for each attention head (see Appendix \ref{apdx:values}).
\subsection{Proposed Method}\label{sec:proposed-methods}

In the following, we drop the head index $i$ for ease of clarity.
Our method addresses cache compression by producing an optimal solution to
$$
\min_{\widetilde{\Kb}}\norm{\Qm \Km^\ts - \Qm\widetilde{\Kb}^\ts}_F^2\ \text{s.t.}\ \mathrm{rank}(\widetilde \Kb) \leq R,
$$
with $\Qm = \begin{pmatrix} \qv^{(j)} \end{pmatrix}_{j \in \text{calibration}}$ the collection of query vectors from the calibration set.
This formulation jointly considers the properties of both $\Kb$ and $\Qb$, while taking into account their interaction through the inner product.
The optimization problem can be restated using a projection matrix as
$$\min_{\Sb}\norm{\Kb\Sb\Qb^\ts-\Kb\Qb^\ts}_F^2~~~~~~\text{s.t.}~~~\mathrm{rank}(\Sb)\leq R,$$
where $\Sb\in\RR^{d \times d}$. To be able to compress the key cache during the attention computation, we write the projection matrix $\Sb$ as a product of two matrices, i.e., $\Sb = \Ab \Bb$ with $\Ab,\Bb\in\RR^{d\times R}$.
The minimization  problem tackled by \kqsvd is thus
\begin{align}\label{eq:min-optimal}
    \min_{\Ab,\Bb\in\RR^{d\times R}}\norm{\Kb\Ab\Bb^\ts\Qb^\ts-\Kb\Qb^\ts}_F^2.
\end{align}
Given $\Ab^*$ and $\Bb^*$ the solutions of this optimization problem, we store the low-rank matrix $\Km \Ab^*\in\RR^{T\times R}$, which allows for compression while maintaining an accurate reconstruction of the attention matrix. The same strategy also applies to the value–output matrices~(see Appendix~\ref{apdx:values}).
In the next section, we state the main theorem which demonstrates that the optimization problem described above admits an optimal solution which can be computed efficiently.

\subsection{\methodname: Optimal attention factorization}
The following theorem establishes the provably optimal low-rank factorization of the key–query matrix, which admits a simple closed-form solution.
\begin{theorem}\label{thm:optimal-kqsvd}
Let $\K,\Q\in \R^{T\times d}$ be key and query cache matrices. The optimal solution to the low rank attention approximation problem
$$\min_{\A,\B\in\R^{d\times R}} \|\K\A\B^\ts\Q^\ts - \K\Q^\ts\|_F,$$
is given by 
$$\A = \K^\pinv\hat{\Ub},\ \ \B = \K^T\hat{\Ub},$$
where $\hat\Ub\in\R^{T\times R}$ is the matrix having the top $R$ left singular vectors of $\K\Q^\top$ as columns. 
\end{theorem}
\begin{proof}

Observe that $\K\A\B^\top\Q^\top$ has rank at most $R$. Hence, 
$$ \min_{\A,\B\in\R^{d\times R}} \|\K\A\B^\ts\Q^\ts - \K\Q^\ts\|_F,$$
is lower bounded by 
$$\min_{\Mb \in \R^{T\times T}} \|\Mb - \K\Q^\ts\|_F \ \text{s.t.}\  \mathrm{rank}(\Mb) \leq R .$$
By the Eckart-Young theorem, we know the best rank $R$ approximation of $\K\Q^\top$ is given by its truncated SVD:
$$\K\Q^\top \simeq \hat{\Ub}\hat{\Sigmab}\hat{\Vb}^\ts,$$
where $\hat{\Ub}\in\RR^{T\times R}, \hat{\Sigmab}\in\RR^{R\times R}$ and $\hat{\Vb}\in\RR^{T\times R}$.
We will show that $\K\A\B^\ts\Q^\ts = \hat{\Ub}\hat{\Sigmab}\hat{\Vb}^\ts $ from which the optimality of $\A$ and $\B$ follows. 

Let $\Kb= \Ub\Sigmab\Vb$ be the full SVD of $\Kb$. 
Observe that we have the inclusion of (column) spans: $\mathcal{R}(\hat \Ub)\subseteq \mathcal{R}(\Ub)= \mathcal{R}(\Kb)$.
Since $\Kb\Kb^{+}$ is the orthogonal projection onto $\mathcal{R}(\Kb)$,  we have $\Kb\Kb^+ \hat \Ub = \hat \Ub$, hence
\begin{align*}
\K\A\B^\ts\Q^\ts  &=
\Kb\Kb^{+}\hat{\Ub}\hat{\Ub}^\ts\Kb\Qb^\ts
= \hat{\Ub}\hat{\Ub}^\ts\Kb\Qb^\ts \\
&= \hat{\Ub}\hat{\Ub}^\ts\Ub\Sigmab\Vb^\ts
= \hat{\Ub}\hat{\Sigmab}\hat{\Vb}^\ts. 
\end{align*}
Therefore $\Ab=\Kb^{+}\hat{\Ub}$ and $\Bb=\Kb^\ts\hat{\Ub}$ are optimal solutions to $\min_{\A,\B\in\R^{d\times R}} \|\K\A\B^\ts\Q^\ts - \K\Q^\ts\|_F$.
\end{proof}

The Moore-Penrose pseudo inverse of $\Kb$ can be expressed through the SVD of $\Kb$ as $\K^{+} = \Vb_K \Sigmab^{-1}_K \Ub_K^\ts$. The singular value decomposition of $\Kb \Qb^\ts \in \R^{T\times T}$ can be computed efficiently as its rank is at most $d$. Indeed, we can first perform an SVD on $\Kb = \Ub_K \Sigmab_K \Vb_K^\ts$ and $\Qb = \Ub_Q \Sigmab_Q \Vb_Q^\ts$, then on the $d \times d$ matrix $\Sigmab_K \Vb_K^\ts \Vb_Q \Sigmab_Q = \Ub' \Sigmab' \Vb'^\ts$. The SVD of $\Kb \Qb^\ts$ is then $\Ub_K \Ub' \Sigmab' (\Vb_Q \Vb')^\ts = \Ub \Sigmab \Vb^\ts$ with $\Ub = \Ub_K \Ub' \in \R^{T\times d}$  and $\Vb = \Vb_Q \Vb' \in \R^{T \times d}$. This way the optimal solution provided by Theorem \ref{thm:optimal-kqsvd} can be computed efficiently in time $\mathcal{O}(T d^2)$. 

A similar approach is used by \methodname to derive optimal projections for the low rank approximation of the product of the value cache with the output matrix~(see Appendix~\ref{apdx:values}).
\section{Theoretical analysis}
In this section, we focus on the minimization problem in Eq.~\eqref{eq:min-optimal}; the same reasoning applies to value and output matrices (see Appendix \ref{apdx:values}). We first provide an exact formula quantifying the accuracy difference between \ksvd~ and \kqsvd. We then exhibit a failure mode of \eigen~method~\citep{saxena2024eigen} which \methodname circumvent by design. Finally, we show how \methodname also obtains optimal low rank approximation in the Grouped Query Attention~(\gqa) framework.
\subsection{Comparing \ksvd and \kqsvd}
In this section, we characterize the optimality gap between \kqsvd and \ksvd~(Section \ref{sec:ksvd}) and derive a closed form expression of their accuracy difference.

\begin{theorem}
Let $\mathrm{opt} = \min_{\A,\B\in\R^{d\times R}} \|\K\A\B^\top\Q^\top - \K\Q^\top\|_F^2$ be the optimal error for the low rank attention approximation problem~(achieved by \methodname), and let $\mathrm{err}_{\text{K-SVD}} = \|\Km \hat{\Vb}_{K} \hat{\Vb}_{K}^T \Qm^T - \Km \Qm^T\|_F^2 $ be the error of~\ksvd. We have
$$\mathrm{err}_{\text{K-SVD}} -  \mathrm{opt} = \sum_{i=1}^R \sigma_i(\Kb\Qb^\top)^2 - \|\Kb \hat{\Vb}_{K} \hat{\Vb}_{K}^T \Qm^T\|^2_F \geq 0,$$
with equality only if the top $R$ left singular vectors of $\Kb$ and the top $R$ of $\Kb\Qb^\top$ span the same subspace.
\end{theorem}
\begin{proof}
Let $\Kb \approx \hat{\Ub}_{K} \hat{\Sigmab}_{K} \hat{\Vb}_{K}^\top$ be the rank $R$ truncated SVD of $\Kb$ and $\Kb\Q^\top \approx \hat{\Ub} \hat{\Sigmab} \hat{\Vb}^\top$ be the one of $\Kb\Qb^\top$.

By Theorem~\ref{thm:optimal-kqsvd}, $\mathrm{opt} = \|\hat{\Ub} \hat{\Ub}^\ts \Kb \Qb^T - \Kb \Qb^T\|_F^2 = \sum_{i=R+1}^d \sigma_i(\Kb \Qb^T)^2$.

On the one hand we have  
\begin{align*}
\|\Kb \Qb^T\|_F^2 = \sum_{i=1}^d \sigma_i(\Kb \Qb^T)^2 
= \sum_{i=1}^R \sigma_i(\Kb \Qb^T)^2 + 
\mathrm{opt}.
\end{align*}

On the other hand,
\begin{align*}
    &\|\Kb \Qb^\ts\|_F^2 
    = \|\Kb \Qb^\ts - \hat{\Ub}_K \hat{\Ub}_K^\ts \Kb \Qb^\ts + \hat{\Ub}_K \hat{\Ub}_K^\ts \Kb \Qb^\ts\|_F^2\\
    &= 
    \|(\Ib - \hat{\Ub}_K \hat{\Ub}_K^\ts) \Kb \Qb^\ts \|_F^2 + \|\hat{\Ub}_K \hat{\Ub}_K^\ts \Kb \Qb^\ts\|_F^2,
\end{align*}
since $(\Ib - \hat{\Ub}_K \hat{\Ub}_K^\ts)$ and $\hat{\Ub}_K \hat{\Ub}_K^\ts$ are projections on orthogonal subspaces. 

Since $\Km \hat{\Vb}_{K} \hat{\Vb}_{K}^\ts  = \hat{\Ub}_K \hat{\Ub}_K^\ts \Kb$, the left term is exactly $\mathrm{err}_{\text{K-SVD}}$ and the second term can be reduced to $\hat{\Ub}_K \hat{\Ub}_K^\ts \Kb \Qb^\ts =  \Kb \hat{\Vb}_K \hat{\Vb}_K^\ts \Qb^\ts $. 

Putting everything together, we get
$$\sum_{i=1}^R \sigma_i(\Kb \Qb^T)^2 + \mathrm{opt} = \mathrm{err}_{\text{K-SVD}} + \| \Kb \hat{\Vb}_K \hat{\Vb}_K^\ts \Qb^\ts \|_F^2,$$
which shows the equality in the theorem. 

To show that   $\mathrm{err}_{\text{K-SVD}} - \mathrm{opt} \geq 0$, first observe that 
$$\mathrm{err}_{\text{K-SVD}} - \mathrm{opt} = \|\hat \Ub \hat \Ub^\top \K\Q^\top \| -  \|\hat \Ub_K \hat \Ub_K^\top \K\Q^\top \|.$$
A direct consequence of the Eckart-Young theorem is that, for any matrix $\Mb\in\R^{T\times T}$ and any $R\leq T$, the solution of 
$$\max_{\Xb\in \R^{T\times R} } \| \Xb\Xb^\ts \Mb \|~~~\text{s.t.}~~\ \Xb^\ts\Xb = \Ib$$
is obtained by setting the columns of $\Xb$ to the top $R$ left singular vectors of $\Mb$. Hence, $\|\hat \Ub \hat \Ub^\top \K\Q^\top \| \geq  \|\hat \Ub_K \hat \Ub_K^\top \K\Q^\top \|$ and thus $\mathrm{err}_{\text{K-SVD}} - \mathrm{opt} \geq 0$, with equality only if $\hat \Ub \hat \Ub^\top= \hat \Ub_K \hat \Ub_K^\top $, i.e., when the top $R$ left singular vectors of $\Kb$ and the top $R$ of $\Kb\Qb^\top$ span the same subspace.
 \end{proof}


It is worth observing that equality between $\mathrm{err}_{\text{K-SVD}}$ and $\mathrm{opt}$ happens \emph{only when} the projection onto $\hat{\Vb}_{K}$ captures \emph{all} of the energy (Frobenius norm) in the top $R$ singular values of $\Kb \Qb^\ts$. Since the best rank-$R$ approximation in the Frobenius norm is unique, this holds precisely when the subspace spanned by the top $R$ left singular vectors of $\Kb$ coincides with that spanned by the top $R$ left singular vectors of $\Kb\Qb^\ts$. In other words, equality holds \emph{precisely when these two subspaces match}.


\subsection{Comparing \eigen~and \kqsvd}

\label{sec:comp_eigen_kq}
We now compare \eigen~\citep{saxena2024eigen} with~\kqsvd. Although we do not derive an exact value for the optimality gap of \eigen, we identify a critical limitation of~\eigen: the method is highly sensitive to unbalance between the norms of $\Kb$ and $\Qb$. 
\eigen's performance can be degraded simply by multiplying $\Kb$ by a constant $\beta$ and dividing $\Qb$ by the same constant. While this rescaling leaves the attention computation unchanged and does not affect \methodname, it causes~\eigen~method to behave increasingly like \ksvd~as the unbalance between the two norms grows.
Theorem~\ref{thm:eigen-vs-kqsvd} formalizes this intuition.

\begin{theorem}\label{thm:eigen-vs-kqsvd}
\label{thm:unbalanced}
Let $\Kb,\Qb\in \RR^{T\times d}$, let $\alpha = \frac{\|\Qb\|_F}{ \|\Kb\|_F}$ and let $\mathrm{err}_{\text{Eigen}} = \norm{\Kb \hat{\Vb}_{\text{Eigen}} \hat{\Vb}_{\text{Eigen}}^\ts\Qb^\ts - \Kb \Qb^\ts} $ be the error of Eigen. If there is a non-trivial gap between the $R$th and $(R+1)$th singular values of $\Kb$, i.e. $\sigma_R(\K) > \sigma_{R+1}(\K)$, then
$\lim_{\alpha \to 0} \mathrm{err}_{\text{Eigen}} = \mathrm{err}_{\text{K-SVD}}$.
\end{theorem}
\begin{proof}
Recall that \eigen approximate $\Kb\Qb^\top$ with $\Kb \hat \V_{Eig} \hat \V_{Eig }^\ts\Qb^\top$ where $\hat \V_{Eig}$ is the matrix with the top $R$ right singular vectors of $\begin{bmatrix}
\Kb \\ 
\Qb
\end{bmatrix}$ as columns.
In the limit where $\frac{\|\Qb\|_F}{ \|\Kb\|_F}$ tends to $0$, the concatenated matrix $\begin{bmatrix}
\Kb \\ 
\Qb
\end{bmatrix}$ tends to  
$\begin{bmatrix}
\Kb \\ 
\mathbf{0}
\end{bmatrix}$. 
Since $\sigma_R(\K) > \sigma_{R+1}(\K)$, it follows from the Davis-Kahan theorem~\citep{davis1970rotation} that, as $\alpha$ tends to $0$, the space spanned by the top $R$ right singular vectors of $\begin{bmatrix}
\Kb \\ 
\Qb
\end{bmatrix}$ converges to the one of $\begin{bmatrix}
\Kb \\ 
\mathbf{0}
\end{bmatrix}$, and thus of $\K$. Hence $\lim_{\alpha \to 0} \hat \V_{Eig}\hat \V_{Eig}^\ts = \hat\V_K\hat\V_K^\ts$ and $\mathrm{err}_{\text{Eigen}}$ tends to $ \mathrm{err}_{\text{K-SVD}}$.
\end{proof}

\subsection{Handling grouped query attention}
Standard multi-head attention (MHA) is powerful but slow at inference, while multi-query attention (MQA)~\citep{shazeer2019fast} is much faster but can hurt model quality and requires retraining. Grouped-query attention (\gqa) \citep{ainslie2023GQA}, sits in between MHA and MQA by letting groups of query heads share a key-value head, balancing accuracy and performance. 
\gqa~organizes attention/query heads into groups of size $m$. All query heads within a group share the same set of keys and values, enabling more efficient computation without significantly compromising model performance. We have assumed in the presentation of our method that each key head attends to a single query head. We show in the following theorem that simply applying \methodname on the shared key cache and the concatenated query caches provides the optimal approximation with \gqa.

\begin{theorem}\label{thm:gqa}
    Given a (shared) key cache matrix $\Kb\in\Rbb^{T\times d}$ and $m$ full column rank query cache matrices $\Qb_1,\cdots,\Qb_m \in \Rbb^{T \times d}$, solving 
    $$\min_{\Ab,\Bb_1,\cdots,\Bb_m\in \R^{d\times R}} \sum_{i=1}^m \| \Kb\Ab\Bb_i^\top\Qb_i^\top - \Kb\Qb_i^\top\|^2_F,$$
    is equivalent to solving
    $$ \min_{\Ab,\Bb\in \R^{d\times R}}  \| \Kb\Ab\Bb^\top\Qb^\top - \Kb\Qb^\top\|^2_F,$$
where $\Qb = [\Qb_1^\top\ \Qb_2^\top\ \cdots\ \Qb_m^\top]^\top \in \mathbb{R}^{mT\times d}$ is the matrix obtained by stacking the $m$ query matrices.
\end{theorem}
\begin{proof}
    We first show that the solution matrices $\Bb_i$ can be chosen to be all equal, i.e., that solving $$\min_{\Ab,\Bb_1,\cdots,\Bb_m\in \R^{d\times R}} \sum_{i=1}^m \| \Kb\Ab\Bb_i^\top\Qb_i^\top - \Kb\Qb_i^\top\|^2_F,$$
    is equivalent to solving 
    $$\min_{\Ab,\Bb\in \R^{d\times R}} \sum_{i=1}^m \| \Kb\Ab\Bb^\top\Qb_i^\top - \Kb\Qb_i^\top\|^2_F. $$
    Indeed, for any $\Ab$ and any $i$, since $\Qb_i^\top$ is full row rank,  the minimizers of $\min_{\Bb_i}\| \Kb\Ab\Bb_i^\top\Qb_i^\top - \Kb\Qb_i^\top\|^2_F$ are the same as the minimizers of $\min_{\Bb_i}\| \Kb\Ab\Bb_i^\top - \Kb\|^2_F$, and are thus independent of $\Qb_i$. Since this is true for any $\Ab$, and in particular for the optimal one, this shows that all the solutions $\Bb_i$ can be taken to be equal. 
    
    The result then directly follows from the fact that the squared Frobenius norm of a block matrix is equal to the sum of the squared Frobenius norms of the blocks. 
\end{proof}

Theorem \ref{thm:gqa} states that optimal projections can be computed for models using GQA by stacking query matrices in each group and using \kqsvd as in the non GQA case. Computing the optimal projection for a head group costs $\mathcal{O}(m T d^2)$, where $m$ is the size of the head group, leading to an amortized cost per query head of $\mathcal{O}(T d^2)$.

\section{Experiments}
\label{sec:exp}
\begin{figure*}[t]
    \centering
    \includegraphics[width=1\linewidth]{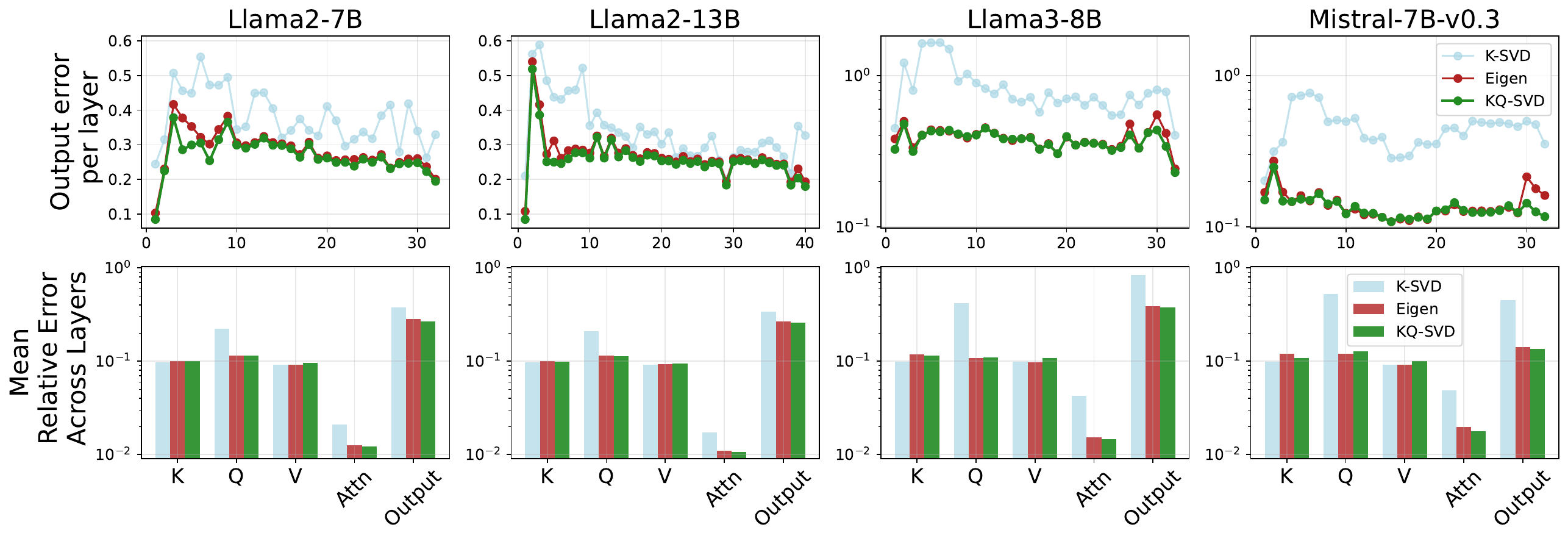}
    \caption{Relative Frobenius approximation output error per layer (top) and mean relative errors on $\Kb$, $\Qb$, $\Vb$, $\Kb\Qb^\ts$ and attention layer output across layers (bottom) for Mistral and Llama models.}
    \label{fig:models_plots}
\end{figure*}
The theoretical results established in the previous sections require further validation on real cache matrices generated by state-of-the-art LLMs. In this section, we first compare \methodname with K-SVD and Eigen, and then demonstrate the claim of Theorem \ref{thm:unbalanced} using real-world cache data.

\subsection{Comparing methods}
\label{sec:comp_mthds}

\textbf{Setup:} To validate our theoretical claims, we run experiments on several widely used LLMs. We test two models without grouped-query attention~(\gqa)—Llama2-7B and Llama2-13B~\citep{touvron2023llama}—and two models with GQA—Llama3-8B~\citep{grattafiori2024llama} and Mistral-7B-v0.3~\citep{jiang2023mistral7b}. All experiments use the C4 dataset~\citep{C4}, with projections learned on the training split and evaluated on the validation split.

\textbf{Learning projections:} Following the methodology of \cite{saxena2024eigen}, we select 128 training sequences of 2048 tokens each from C4. Each sequence is passed through the model, storing the key, value, and query caches for every layer and attention head (queries are needed by both \eigen~and \kqsvd). For each layer–head pair, we collect the corresponding caches from all 128 sequences and concatenate them, yielding large matrices $\Km, \Qm, \Vm \in \R^{T_{\text{huge}} \times d}$ with $T_{\text{huge}} = 262{,}144$.
Because model context length is limited and attention cost scales quadratically with sequence length, it is not feasible to build these matrices from a single long sequence. Instead, we construct them from multiple shorter ones. Once the large cache matrices are formed, we perform SVD and apply the formulas presented in previous sections to compute low-rank projections.

\textbf{Rank selection:} All methods are evaluated at the same rank $R$, determined individually for each layer. For a given layer, we analyze the singular value spectra of the key and value matrices, averaged across heads, and choose the smallest $R$ that discards no more than an $\epsilon = 0.1$ fraction of the spectral energy. That is, for singular values $\{\sigma_j\}_j$ of $\Km$,
$\frac{\sum_{j=1}^{R} \sigma_j^2}{\sum_{i=1}^d \sigma_j^2} \ge 1 -\epsilon$.
This is equivalent to requiring that the relative Frobenius error is at most $\epsilon$.

\textbf{Evaluation:} We evaluate the learned projections on 32 validation sequences of 2048 tokens each. For every sequence, we extract the cache matrices ($\Km, \Qm, \Vm$) at each layer and head. Using these matrices, we can simulate attention computations directly, since attention depends only on these three components. Each cache is projected onto its corresponding low-rank subspace to form approximations, and we then compute the approximate Multi-Head Attention output using the standard formulas. For comparison, we also compute the exact (uncompressed) attention output. This enables us to measure the relative error of each method across all matrices of the attention pipeline. Errors are averaged across validation sequences.

\textbf{Metrics:} For a matrix $\Mb$ and its approximation $\widetilde{\Mb}$, we report the relative Frobenius norm error
$\text{err}_{\text{Fro}} = \frac{\|\Mb-\widetilde{\Mb}\|_F^2}{\|\Mb\|_F^2}$.
This error is computed for the key, query, and value matrices, for the attention score matrix $\Kb \Qb^\ts$, and for the Multi-Head Attention output $\mathrm{MHA}(\Xb)$.

\textbf{Results:} Results are shown in Figure~\ref{fig:models_plots}. 
For each model, the top plot reports the relative error on the attention output across layers, while the bottom plot reports the averaged errors on the intermediate components. We observe that \ksvd~provides the most accurate approximation of the key matrices (as expected from the optimality of SVD), but performs poorly on query matrices, leading to weaker approximations of the attention scores and consequently higher output errors. This effect is more pronounced in \gqa~models, where sharing the key matrix across queries in a group amplifies approximation errors.

In contrast, \eigen~and \kqsvd~achieve comparable accuracy on keys, queries, and values. The key difference lies in the attention score matrix: \kqsvd~consistently delivers higher accuracy, resulting in lower  attention errors and outperforming all other methods on all models.

\subsection{Unbalanced $\Km$ and $\Qm$ matrices}

We verify experimentally the claims of Section \ref{sec:comp_eigen_kq} and analyze the attention approximation error under unbalanced $\Km$ and $\Qm$ matrices.

\textbf{Set up}: We follow the same experimental setup as in the previous section using the C4 train/validation split to learn projections and evaluate their accuracy. The only difference is that cache matrices are scaled to assess the effect of unbalance. Key matrices are multiplied by $\beta$ and query matrices are divided by $\beta$. This is equivalent to scaling the projection matrices $\Wm_{K,i}$/ $\Wm_{Q,i}$ by $\beta$, as the operation permutes with matrix multiplication~(this does not change the output since the query and key matrices are multiplied before any non-linear activation). 


\textbf{Metrics}: For each unbalance ratio, we plot the relative attention output error for the three methods, averaged across all layers and validation sequences. 

\begin{figure}
    \centering
    \includegraphics[width=1\linewidth]{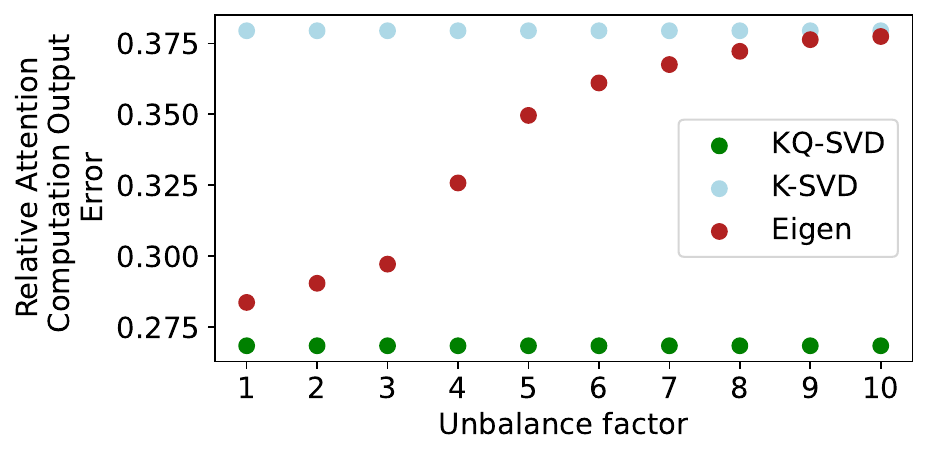}
    \caption{Llama2-7B relative output approximation error averaged across layers for varying unbalanced factor $\beta$.}
    \label{fig:Llama2_7B_unbalanced}
\end{figure}

\textbf{Results}: Results are shown in Figure \ref{fig:Llama2_7B_unbalanced}. As discussed in Section \ref{sec:comp_eigen_kq},  \ksvd~and \kqsvd~are invariant to scaling $\Km$ by a factor $\beta$ and dividing $\Qm$ by the same factor, which is confirmed by the constant error observed for these methods. As predicted theoretically, a higher unbalance ratio brings \eigen~closer to \ksvd; for $\beta=10$, their errors are nearly indistinguishable. This confirms Theorem \ref{thm:unbalanced} and exposes a limitation of \eigen, which underperforms even under modest query-key unbalance.

\section{Conclusion}

We introduced \methodname{}, a novel approach for computing low-rank projections for KV cache compression. \methodname is driven by minimizing an upper bound on the attention output approximation error, resulting in an optimization problem that yields the optimal low-rank approximation of the attention matrix. Crucially, this problem admits a closed-form solution that can be computed efficiently. We quantify the advantage of \methodname{} over previous methods, either through exact error formulas or by highlighting failure modes of prior approaches. Our technique is complementary to popular cache compression methods such as~\gqa. 
Experiments validate our theoretical findings and demonstrate that considering the interaction between queries and keys—as \methodname{}—provides a superior alternative to performing SVD on the key cache or on concatenated keys and queries.
\subsection*{Acknowledgment}
This research is supported by the Canadian Institute
for Advanced Research (CIFAR AI chair program). This work was completed while Damien Lesens interned at Mila. This work made use of compute resources provided by the Digital Research Alliance of Canada and  by Mila (mila.quebec).


\bibliography{bibtex}

\clearpage
\onecolumn

\appendix
\section*{\centering KQ-SVD: Compressing the KV Cache with Provable Guarantees on Attention Fidelity \\ \vspace*{0.3cm}(Supplementary Material)}
\section{Proof of Theorem \ref{thm:kq-bound}}
\label{apdx:thm1}

\begin{theorem*}
     Let $\Xb\in\RR^{T\times d}$ be a sequence of token embeddings, $\Kb,\Qb, \Vb\in\RR^{T\times d}$ and 
     $$\mathrm{MHA}(\Xb) = \big[\mathrm{Softmax}(\Qb_i \Kb_i^\ts / \sqrt{d}) \Vb_i)\big]_i\Wb^{O},$$
    $$\widetilde{\mathrm{MHA}}(\Xb) = \big[\mathrm{Softmax}(\Qb_i \widetilde{\Kb_i}^\ts / \sqrt{d}) \widetilde{\Vb_i})\big]_i\Wb^{O},$$
where $\widetilde{\mathrm{MHA}}(\Xb), \widetilde{\Kb_i}$ and $\widetilde{\Vb_i}$ represent the approximation of $\mathrm{MHA}(\Xb),\Kb_i$ and $\Vb_i$, respectively. The difference between the actual attention output and the one produced with approximate keys and values is upper bounded as
\begin{align}\label{eq:purturbation-bound}
    &\norm{\widetilde{\mathrm{MHA}}(\Xm) - \mathrm{MHA}(\Xm)}_2\nonumber\\
    &\le 
    \sum_{i=1}^{h} \frac{\norm{\Vb_i \Wb_{i}^O}_2}{\sqrt{d}}\norm{\Qb_i \Kb_i^\ts - \Qb_i\widetilde{\Kb_i}^\ts}_2\nonumber\\
    &+ 
    \norm{\Vb_i\Wb_{i}^O-\widetilde{\Vb_i}\Wb_{i}^O }_2.
\end{align}
\end{theorem*}

\begin{proof}
Let $\Wb^O = [\Wb^{O}_1;\Wb^{O}_2;\cdots;\Wb^{O}_h]\in\RR^{D\times D}$, where $\Wb^{O}_i\in\RR^{d\times D}$ are stacked vertically. 
By the definition of multi-head attention (see Section~\ref{sec:background}), we can write
\[
\mathrm{MHA}(\Xb) = [\Hb_1,\dots, \Hb_{h}] \Wb^O = \sum_{i=1}^{h} \Hb_i \Wb^{O}_i, 
\quad 
\widetilde{\mathrm{MHA}}(\Xb) = [\widetilde{\Hb}_1,\dots, \widetilde{\Hb}_h] \Wb^O = \sum_{i=1}^{h} \widetilde{\Hb}_i \Wb^{O}_i.
\]
Therefore,
$$
\norm{ \widetilde{\mathrm{MHA}}(\Xb) - \mathrm{MHA}(\Xb) }_2 
= 
\left\| \sum_{i=1}^{h} (\widetilde{\Hb}_i - \Hb_i)\Wb^{O}_i \right\|_2
\le 
\sum_{i=1}^{h} \| (\widetilde{\Hb}_i - \Hb_i) \Wb^{O}_i \|_2.
$$

For each head $i \in \{1,\ldots,h\}$,
\begin{align}
    \| (\widetilde{\Hb}_i - \Hb_i)\Wb^{O}_i \|_2 
    &= 
    \big\| 
    \big( \Softmax(\tfrac{\Qb_i \Kb_i^\top}{\sqrt{d}})\Vb_i 
    - 
    \Softmax(\tfrac{\Qb_i \widetilde{\Kb}_i^\top}{\sqrt{d}})\widetilde{\Vb}_i \big)
    \Wb^O_i
    \big\|_2 \nonumber\\[2pt]
    &\le
    \big\| 
    \big(\Softmax(\tfrac{\Qb_i \Kb_i^\top}{\sqrt{d}}) - \Softmax(\tfrac{\Qb_i \widetilde{\Kb}_i^\top}{\sqrt{d}})\big)\Vb_i \Wb^O_i 
    \big\|_2
    +
    \big\|
    \Softmax(\tfrac{\Qb_i \widetilde{\Kb}_i^\top}{\sqrt{d}})(\Vb_i-\widetilde{\Vb}_i)\Wb^O_i
    \big\|_2.
    \label{eq:error}
\end{align}

For the first term in~\eqref{eq:error}, applying the submultiplicative property of the 2-norm gives
$$
\big\| 
\big(\Softmax(\tfrac{\Qb_i \Kb_i^\top}{\sqrt{d}}) - \Softmax(\tfrac{\Qb_i \widetilde{\Kb}_i^\top}{\sqrt{d}})\big)
\Vb_i \Wb^O_i 
\big\|_2
\le 
\norm{ \Softmax(\tfrac{\Qb_i \Kb_i^\top}{\sqrt{d}}) - \Softmax(\tfrac{\Qb_i \widetilde{\Kb}_i^\top}{\sqrt{d}}) }_2
\, \norm{ \Vb_i \Wb^O_i }_2.
$$
The factor $\norm{\Vb_i \Wb^O_i}_2$ acts as an amplification term, capturing how sensitivity in the value projections may magnify small perturbations in the attention weights—this term is typically not directly controllable in practice.

Since the Softmax function is $\tfrac{1}{2}$-Lipschitz continuous~(see Appendix A.4 \citep{alghamdi2022beyond}), we have
$$
\norm{ \Softmax(\tfrac{\Qb_i \Kb_i^\top}{\sqrt{d}}) - \Softmax(\tfrac{\Qb_i \widetilde{\Kb}_i^\top}{\sqrt{d}}) }_2
\le 
\frac{1}{\sqrt{d}} \norm{ \Qb_i \Kb_i^\top - \Qb_i \widetilde{\Kb}_i^\top }_2.
$$

For the second term in~\eqref{eq:error}, note that $\norm{ \Softmax(\cdot) }_1 = 1$, implying $\norm{ \Softmax(\cdot) }_2 \le 1$. 
Therefore,
$$
\big\|
\Softmax(\tfrac{\Qb_i \widetilde{\Kb}_i^\top}{\sqrt{d}})(\Vb_i - \widetilde{\Vb}_i)\Wb^O_i
\big\|_2
\le 
\norm{\Vb_i\Wb^O_i - \widetilde{\Vb}_i\Wb^O_i }_2.
$$

Combining the two bounds yields
$$
\norm{ (\widetilde{\Hb}_i - \Hb_i)\Wb^{O}_i }_2
\le
\frac{\norm{\Vb_i \Wb^O_i}_2}{\sqrt{d}} 
\norm{ \Qb_i \Kb_i^\top - \Qb_i \widetilde{\Kb}_i^\top }_2
+
\norm{\Vb_i\Wb^O_i- \widetilde{\Vb}_i\Wb^O_i}_2.
$$
Summing over all heads $i=1,\ldots,h$ gives the desired bound~\eqref{eq:purturbation-bound}, completing the proof.
\end{proof}

\section{Value-Output Projection}
\label{apdx:values}

In this section, we examine the interaction between the value representations and the output projection. 
While the main analysis of this paper focuses on the relationship between keys and queries, the same reasoning naturally extends to values $\Vb_i\in\RR^{T\times d}$ and the output matrix~$\Wb_i^O\in\RR^{d\times D}$. 
To further tighten the upper bound established in Theorem~\ref{thm:kq-bound}, we aim to minimize the second term in the summation, which leads to the following optimization problem:
$$
\min_{\widetilde{\Vb}} \, \big\| \Vb \Wb^{O} - \widetilde{\Vb} \Wb^{O} \big\|_F^2~~~~ \text{s.t.}~~~\mathrm{rank}(\widetilde{\Vb}) \leq R,
$$ 
where for simplicity we drop the subscript.
Likewise for the keys and queries case, the optimization problem can be restated using a projection matrix $\Sb\in\RR^{d\times d}$ such that
$$\min_{\Sb}\norm{\Vb\Sb\Wb^O-\Vb\Wb^O}_F^2~~~~~~\text{s.t.}~~~\mathrm{rank}(\Sb)\leq R,$$
where $\Sb\in\RR^{d \times d}$ where we write the projection matrix $\Sb$ as a product of two matrices, i.e., $\Sb = \Ab \Bb^\top$ with $\Ab,\Bb\in\RR^{d\times R}$.
The minimization problem tackled by \kqsvd for the value and output projections is
\begin{align}\label{eq:min-optimal}
    \min_{\Ab,\Bb\in\RR^{d\times R}}\norm{\Vb\Ab\Bb^\ts\Wb^O-\Vb\Wb^O}_F^2.
\end{align}

Theorem \ref{thm:optimal-kqsvd} applies, in a similar fashion to the case of values and outputs. 

\section{Practical settings}

\textbf{Code:} The code used for experiments presented in Section \ref{sec:exp} is available at \url{https://github.com/DamienLesens/KQ-SVD}. 
We used Pytorch \citep{pytorch} and the Hugging Face \texttt{transformers} library \citep{wolf2020huggingfacestransformersstateoftheartnatural}. For reproducibility, we fixed random seeds equals to 0.

\textbf{Hardware:} All experiments were conducted on NVIDIA Tesla V100-SXM2-32GB GPUs with CUDA acceleration. The primary compute nodes were Intel Xeon E5-2698 v4 @ 2.20GHz (503GB RAM).

\thispagestyle{empty}














\end{document}